\documentclass[preprint]{elsarticle}

\usepackage{lineno,hyperref}
\usepackage{url}
\usepackage{graphicx}
\usepackage{epstopdf}
\usepackage{amsfonts}
\usepackage{amsmath}
\usepackage{bm}
\usepackage{mathtools}
\usepackage{subfig}
\usepackage{makecell}
\usepackage{hyperref}
\usepackage{multirow}
\usepackage{svg}

\DeclareMathOperator*{\argmax}{arg\,max}

\journal{Journal of Artificial Intelligence (AIJ)}

\begin{document}

\begin{frontmatter}

\title{Explaining Predictions of Deep Neural Classifier via Activation Analysis}
% Explaining predictions of deep neural classifier by analysing activations
% Explaining predictions of deep neural classifier via perceptual analysis
% Analysing neural activations to explain predictions of deep neural classifier for medical diagnosis


\author[stua]{Martin Stano \corref{mycorrespondingauthor}}
\cortext[mycorrespondingauthor]{Corresponding author}
\ead{xstano@stuba.sk}

\author[stub]{Wanda Benesova}
\ead{wanda.benesova@gmail.com}

\author[cpjku,litjku]{Lukáš Samuel Marták}
\ead{lukas.martak@jku.at}

\address[stua]{Department of Informatics, Informational Systems and Software Engineering, Faculty of Informatics and Information Technologies, Slovak University of Technology, 842 16 Bratislava, Slovakia}
\address[stub]{Department of Computer Engineering and Applied Informatics, Faculty of Informatics and Information Technologies, Slovak University of Technology, 842 16 Bratislava, Slovakia}
%\address[jku]{Institute of Computational Perception and LIT Artificial Intelligence Lab, Johannes Kepler University, 4040 Linz, Austria}
\address[cpjku]{Institute of Computational Perception, Johannes Kepler University, 4040 Linz, Austria}
\address[litjku]{LIT Artificial Intelligence Lab, Johannes Kepler University, 4040 Linz, Austria}

\begin{abstract}
In many practical applications, deep neural networks have been typically deployed to operate as a black box predictors. Despite high amount of work on interpretability and high demand on reliability of these systems, they typically still have to include a human actor in the loop, to validate the decisions and handle unpredictable failures and unexpected corner cases. This is true in particular for failure-critical application domains, such as medical diagnosis.

We present a novel approach to explain and support an interpretation of a decision-making process to a human expert operating a deep learning system based on Convolutional Neural Network (CNN).

By modeling activation statistics on selected layers of a trained CNN via Gaussian Mixture Models (GMM), we develop a novel perceptual code in binary vector space that describes how input sample is processed by the CNN. By measuring distances between pairs of samples in this perceptual encoding space, for any new input sample, we can now retrieve a set of most perceptually similar and dissimilar samples from an existing atlas of labeled samples, to support and clarify the decision made by the CNN model.

Possible uses of this approach include for example Computer-Aided Diagnosis (CAD) systems working with medical imaging data, such as Magnetic Resonance Imaging (MRI) or Computed Tomography (CT) scans.  We demonstrate the viability of our method in the domain of medical imaging for patient condition diagnosis, as proposed decision explanation method via similar ground truth domain examples (e.g. from existing diagnosis archives) will be interpretable by the operating medical personnel.

%Our results indicate that our method is capable of detecting distinct prediction strategies and potentially identify model over-generalization or other deficiencies.

Our results indicate that our method is capable of detecting distinct prediction strategies which enables to identify most similar predictions from an existing atlas.
\end{abstract}


\end{frontmatter}

%\linenumbers

\section{Introduction}

Usage of deep Convolutional Neural Networks (CNN) in medical image processing has proven to be a successful but hardly interpretable and explainable method.

The problem of interpratability in the medical domain  has been addressed by the authors in the paper \cite{caruana2015intelligible}. They have achieved state-of-the-art accuracy in the prediction of Pneumonia Risk, but also have concluded that although the models accurately explain the predictions they make, they are still based on correlation. It is not clear if some phenomena are caused by over-fitting, correlation with other variables, interaction with other variables, correlation or interaction with unmeasured variables or some additional underlying phenomena.

In fields such as health care, critical decisions can't be based on black-box predictions \cite{holzinger2017we}. In order to get medical experts to trust and use the artificial intelligence diagnosis systems, the decisions of these systems must be explained and interpreted in a way that enables easy verification with an expert's domain knowledge. These explanations and interpretations can then be used to identify shortages of the empirical distribution or flaws of the model architecture itself.

Classical deep neural network "perceives" each input via a set of neuron activations computed as part of the model inference. Complex models designed to process large volumetric medical images, such as CT or MRI scans, consist of hundreds of thousands of neurons. Direct numerical comparison of these activation patterns would be computationally inefficient, given the large dimensionality of activation spaces of deep neural network models. 

We propose a novel and efficient approach for explaining a deep neural network decision process, that is based on a "perceptual" code inferred from the network's intermediate computations of neuron activations. For an explanation of a particular decision process, the "perceptually" closest atlas sample is retrieved and presented to the user along with the prediction. In other words, matching criteria for a retrieval from the atlas follow from similarities in the neural activity signals given by the forward passes of the deep neural network classifier at hand.
Our main contribution breaks down into following parts:
\begin{itemize}
\item encoding of neuron signal activities using Gaussian Mixture Model (GMM) into novel perceptual code
\item method of atlas creation: a set of representative samples to support explanations of future predictions
\item matching process based on the novel perceptual code
\end{itemize}
%In this paper we propose a novel and efficient approach to the atlas creation process and prediction basis retrieval using GMM modeling of activity of each neuron on selected network levels to efficiently compare new data samples with the samples in our atlas. 

By analyzing the activation values of a neuron when processing several images of the same class, we can model the behavior of this neuron using the Gaussian Mixture Model (GMM) and assert if activation value of unknown image complies with this model. This concept allows us to computationally and memory efficiently encode and compare new evaluated data with the samples provided by the atlas.  

Our method helps only with interpretation and explanation of the trained neural network model's decision without altering or constraining the decision or the training process.

%Providing a set of annotated atlas images from the class of the unknown image being processed by the neural network which are similar to the given unknown image in a way the network perceives them may prove useful for the verification of the system and serves as a natural input for the operating medical expert.

\section{Related work}\label{sec:what}

Authors of \cite{samek2017} divide the needs of an explainable AI into four categories: verification of the system's decisions, system improvement by identifying its weaknesses, learning patterns from the system yet unobserved by humans and compliance to strict legislation. These authors also distinguish between terms of \textit{interpretation} and \textit{explanation} defining interpretation as mapping of an abstract concept (an output class) into a domain example (e.g. an image) and explanation as a set of domain features (e.g. image pixels) contributing to the output model decision. 

In their more recent work~\cite{samek2019explainable}, authors also established a conceptual classification of explainability methods into four main categories: (i) \textit{explaining learned representations}, (ii) \textit{explaining individual predictions}, (iii) \textit{explaining model behavior}, and (iv) \textit{explaining with representative examples}.

They define explaining learned representations as a type of explanation that aims to foster the understanding of the learned representations, e.g., neurons of a deep neural network by \textit{generating prototypical outputs} representing classified classes (e.g. images of cars for class "car").

Next, they define explaining individual predictions as a group of methods visualizing which \textit{parts of a single input} have been most relevant for the model to arrive at its decision, e.g, in a form of a heat map.

Thirdly, explaining model behavior aims to go beyond the analysis of individual predictions towards a more general understanding of model behavior, e.g., \textit{identification of distinct prediction strategies}.

Lastly, explaining with representative examples interpret classifiers by identifying representative training examples. This type of explanations can be useful for obtaining a better understanding of the training dataset and how it influences the model.

The extent to which an explanation of a statement to a human expert achieves a specified level of causal understanding with effectiveness, efficiency and satisfaction in a specified context of use is defined as \textit{causability} \cite{london2019artificial}. The concept of causability in human-AI interaction is analogous to usability in classic human-computer interaction. To measure the causability of an explanation of a machine statement this has to be based on a causal model, which is not the case for most ML algorithms, thus a mapping between both has to be defined (also called \textit{explanation interface}).

Several methods for explaining and interpreting deep neural networks have already been devised, and here we will briefly review some of them.

Activation maximization is a model interpretation technique which searches for input domain sample (prototype) $\mathbf{x}^{\ast}$ which produces maximal response (neuron activation) for given output class $\omega_c$ \cite{erhan2009visualizing}. Formally:

\begin{equation}
    \mathbf{x}^{\ast} = \argmax_{\mathbf{x}} \log p(w_c|\mathbf{x}) - \lambda {\|\mathbf{x}\|}^2
    \label{eq:am}
\end{equation}

where the conditional probability $p(w_c|\mathbf{x})$ denotes the activation value of corresponding output neuron with softmax activation. Since this term is modeled by the neural network it is guaranteed to have a gradient, therefore this optimization problem can be solved using gradient methods.

Sensitivity analysis addresses the problem of explainability. It assigns a relevancy score to each input feature of the model, thus identifying the most important features for the task at hand (e.g. classification or regression). In the case of visual images, a score for each pixel of the image. Sensitivity analysis can be seamlessly implemented in deep neural networks as an extension of the backpropagation algorithm, thus sometimes called backpropagation to data. In the case of image classification, the relevancy scores assigned to input pixel can be visualized in the form of a heatmap.

Results of simple sensitivity analysis can be scattered and discontinuous for even the simplest benchmark image datasets. More robust approaches such as the Layer-wise Relevance Propagation (LRP) \cite{bach2015pixel} have thus been explored. LRP leverages the graph structure of deep neural networks. The key concept of this method is a layer to layer redistribution of the initial relevancy given by the network output backwards from the output layer up to the input data while maintaining the \textit{conservation property}, which is defined as follows: the sum of all relevancy shares received by a neuron from the preceding layer equals to the sum of all relevancy shares redistributed to the succeeding layer.

A combination of interpretation and explanation has been successfully used in \cite{lee2018haemorrhage} for acute intracranial hemorrhage (ICH) diagnosis from CT images. As an explanation technique, the authors used class activation mapping (CAM) to localize and visualize the areas of hemorrhage. To further support and interpret each decision of the network, they also provided several activation-wise similar images (called prediction basis) from custom composed atlas consisting of selected training examples. This particular approach of exposing the activation-wise similar prediction-basis is also used in~\cite{Tamajka2019}.

With respect to the classification of explainability methods given by \cite{samek2019explainable}, our method is a combination of \textit{explaining model behavior} and \textit{explaining with representative examples} while also complying with parts of the definition of \textit{explaining learned representations} (i.e. understanding learned representations of neurons). First, we create an atlas based on learned representations (the activity of neurons). The structure (clustering) of the created atlas helps us to explain distinct activation patterns for the intra-classes of a given class. Lastly, prediction is explained visually with representative examples retrieved from our atlas.

The idea of clustering model-behavioral meta-information about individual data points is also exploited by the spectral relevance analysis (SpRAy) approach \cite{lapuschkin2019unmasking} which identifies distinct prediction strategies by clustering individual heatmaps.

Our approach also relates to the framework of prototypical learning~\cite{Bien2011}, where prototype selection is formulated as an optimization problem with an emphasis on capturing the intra-class diversity of the training samples. More recently, various methods for prediction basis (also called "prototypical samples") retrieval have been devised~\cite{Koh2017, Yeh2018, Li2017}.

Using influence functions~\cite{Koh2017}, one can trace a model's prediction all the way back to the training data, identifying training points most responsible for a given prediction. Furthermore, prototypes can be selected from the training set based on perceptual similarity metrics, such that activation of a test sample can be represented by a linear combination of activations of these prototypes, yielding weights that correspond to the importance of individual prototypes, denoted as \emph{representer points}~\cite{Yeh2018}. 

Alternatively, prototypes can also be \emph{learned} in the latent space of a deep neural network and encoded there in form of a \emph{prototype layer}~\cite{Li2017}. This approach allows for a flexible determination of prototypes in latent space, but their closeness to some training point as well as the quality of their reconstruction back to the data space both depend on the success of learning, which might not be the optimal approach for medical imaging applications, where domain experts that would end up examining the prediction basis expect detailed prototypes from a database of real data points to lean on.

One other problem that is closely related to our method is the detection of Out-Of-Distribution (OOD) samples. Authors of~\cite{Shafaei2018} recognize four paradigms in previous works on related problems: based on (i) uncertainty quantification, (ii) abstention from prediction, (iii) anomaly detection, and (iv) novelty detection. They further distinguish more recent deep learning approaches based on whether In-Distribution or Out-Of-Distribution samples are assumed to be on the input during test time and propose a novel methodology for a comprehensive evaluation of outlier detectors they call OD-test. Even though we don't target the OOD sample detection problem in this contribution, our approach fits the anomaly detection paradigm, and could be easily extended to perform OOD sample detection, and further OD-tested accordingly.

\section{Proposed method} \label{proposal}

\begin{figure}[h]
    \centering
    \includegraphics[width=\columnwidth]{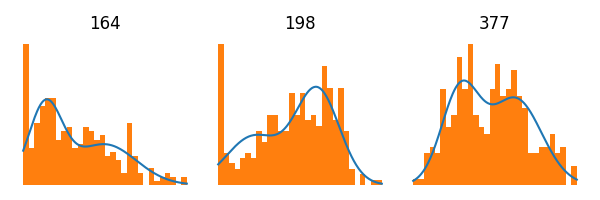}
    \caption{Example of GMM approximated histograms for three specific neurons}
    \label{fig:hist}
\end{figure}

Our proposed explainability method stems from providing prediction basis (similar atlas images) to novel input samples. It consists of four main stages: 

\begin{enumerate}
    \item \textbf{Creation of histograms} of artificial neuron activity for each neuron on selected layers of the network. Analyzed activations are caused by passing all training images of a particular class forward through a trained network.
    \item \textbf{Gaussian mixture model (GMM) approximation} of created histograms. Histogram of each neuron is modeled as a linear combination of two Gaussian mixture components, as we observed at most two modes when examining histograms (example in Fig. \ref{fig:hist}).
%elaborativna diskusia na Skype:
% * lebo vacsina histogramov na prvy pohlad mala max 2 mody
% * po dalsom preskumani sme zistili ze vacsinou iba 1 mod
% * skusime celu metodu s 1 modom konstantne
% * je mozne skusit v buducnosti napr aj dynamicky pocet modov pri skusani napr. 1-5 s goodness of fit
% * je mozne zapojit do aproximacie a modelu aj skewness parameter pre jednotlive mody pre zvyseny goodness of fit
    \item \textbf{Encoding of an image} based on affiliation to approximated GMM components. Once a novel sample passed through the network, class prediction is computed, and activations of selected subset of neurons are stored. Subsequently, each activation is evaluated under its corresponding GMM. This results in a binary encoding that indicates affiliation of each activation value to individual components of its GMM (see Fig. \ref{fig:encoding} for an example with 2 GMM components). For a choice of $T$ Gaussians per GMM and a subset of $M$ neurons to keep track of, this yields a binary perceptual code of dimension $M T$.
    \item \textbf{Atlas creation and prediction basis retrieval}. To create an atlas, stage 3 is repeated for each sample in a set of known, labeled inputs. It is possible to use the training data set. Prediction basis consists of $k$ most similar images from the atlas determined by picking $k$ samples from the atlas with smallest Hamming distances to the new input, as measured on their binary perceptual codes.
\end{enumerate}

% PLACEHOLDER NA FIGURE - HISTOGRAMY

\begin{figure}[h]
    \centering
    \includegraphics[width=\columnwidth]{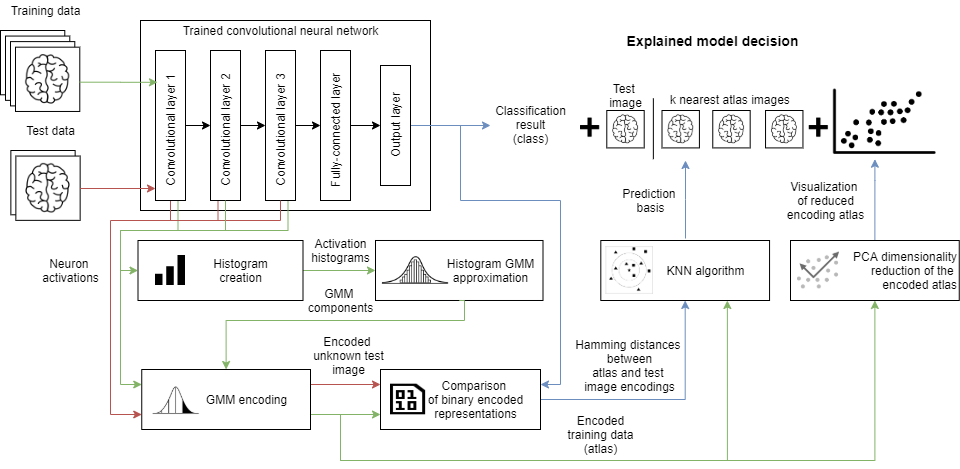}
    \caption{Overview of our proposed method.}
    \label{fig:methodOverview}
\end{figure}

A schematic graphic overview of our explanation method's pipeline showing the relations and data flow between the stages is depicted in Fig. \ref{fig:methodOverview}

In the remainder of this section, we describe these four stages of our method in further detail.

\subsection{Creation of histograms} \label{sub::histo}

Lets consider a trained 3D CNN classifier model classifying into $\left | C \right |$ classes which consists of multiple stacked convolutional and pooling layers. When processing input 3D image volume $\mathbf{I}$, each of the layers $l \in L$ produces an output activation volume $\mathbf{V}_{l+1} = F_{l}(\mathbf{V}_{l})$ where $F_l$ denotes both the operation (convolution or pooling) of corresponding layer $l$ and application of its activation function. Each activation volume $\mathbf{V}_l$ can then be serialized into a vector $\mathbf{u}_l$ of layer-specific dimensionality $WHD$ where $W$, $H$ and $D$ denote the width, height and depth specific for $l$-th activation volume $\mathbf{V}_l$.

Passing $K$ inputs of class $c \in C$ through the network produces an activation matrix $\mathbf{A}^{(c)} \in \mathbb{R}^{K \times M}$ for all $M$ neurons in subset of layers $L' \subseteq L$ that we pick to keep track of, where 
\begin{equation}
    M = \sum_{l \in L'} | \mathbf{u}_l |.
\end{equation}

We create $M$ histograms $\mathbf{h}^{(c)}_{j} \in \mathbb{R}^{B}$, one for each of $j \in \{1, \dots, M\}$ neurons where $B$ is the chosen number of bins in the histogram. This histogram represents frequency of activation values of neuron $j$ while the network processed the $K$ given images. 

\subsection{Distribution approximation}

Each histogram $\mathbf{h}^{(c)}_{j}$ can then be approximated by a Gaussian Mixture Model (GMM) defined as:

\begin{equation}\label{eq:mog}
p(x) = \sum_{t = 1}^{T}{w_t}\mathcal{N}(x; \mu_t, \sigma_t^2)
\end{equation}

where $T$ is the number of mixture components, $\mathcal{N}$ is normal probability distribution function with mean $\mu_t$, variance $\sigma_t^2$ and weight of $t$-th component $w_t$. Through this approximation, histogram representation is reduced to the triplet of $(w_t, \mu_t, \sigma_t^2)$ values for all $T$ components; a matrix $\mathbf{H}^{(c)}_{j} \in \mathbb{R}^{T \times 3}$. Since two GMM components are used in our experiments, each histogram can be stored using $6$ values. We applied Expectation-Maximization (EM) algorithm~\cite{dempster1977maximum} to fit the parameters of each GMM to the approximated histogram data.

\subsection{Encoding of an image} \label{sub::encoding}

Given our approximation of activation histograms, an image $\mathbf{I}$ processed by the network can also be memory efficiently encoded as a binary vector representing affiliations of neural activations to components of respective distributions, as approximated by our GMM models. Consider activation value $v$ of neuron $j$ computed from input $\mathbf{I}$ of class $c$. To encode $v$, which is affiliated to histogram approximation $\mathbf{H}^{(c)}_{j}$, we define  $\mathbf{e}_v = e_{v1} \oplus e_{v2} \oplus \dots \oplus e_{vT}$, where operator $\oplus$ denotes the operation of concatenation and $e_{vt}$ denotes the affiliation of $v$ to $t$-th GMM component of $\mathbf{H}^{(c)}_{j}$.
Truth value of $e_{vt}$ is defined as follows:

\begin{equation}\label{eq:encoding}
\begin{aligned}
e_{vt} = 
\begin{cases}
    1   & \quad \text{if } (\mu_{t} - \sigma^2_t \leq v \leq \mu_{t} + \sigma^2_t) \\ 
    & \quad \wedge  \quad \text{component $t$ is relevant} \\
    0   & \quad \text{otherwise}
\end{cases}
\end{aligned}
\end{equation}

Criteria of component relevancy are subject to further research and validation. Our experiments were conducted with relevancy of $t$-th component determined by rule

\begin{equation}
    \mathcal{N}(\mu_t; \mu_t, \sigma_t^2) > q * \frac{1}{N}\sum^{N}_i{\mathcal{N}(\mu_i; \mu_i, \sigma_i^2)},
\end{equation}

where $\frac{1}{N}\sum^{N}_i{\mathcal{N}(\mu_i; \mu_i, \sigma_i^2)}$ is mean of probability peak values across all GMM components of all activations, thus $N = MT$. Scaling constant $q$ is set as a hyper-parameter. Intuitively, a component is relevant if its probability peak is above $q$-scaled average of probability peaks of all components of neural activations. Other alternatives based on the high variance of component or component weight might be also considered.

\begin{figure}[htbp]
    \centering
    \includegraphics[width=0.7\linewidth]{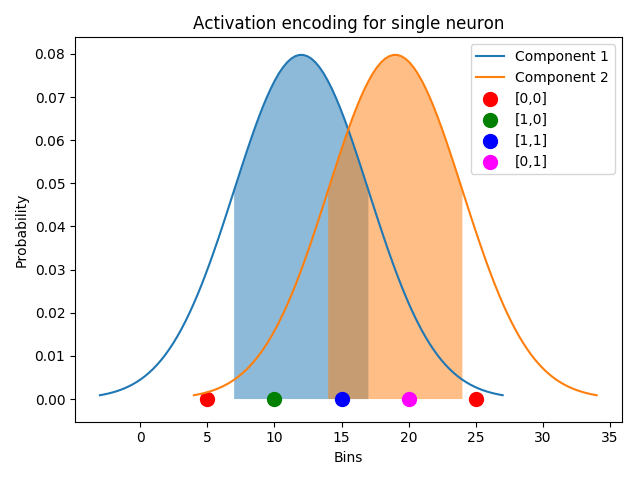}
    \caption{Example of activation encoding for single neuron. The two Gaussian curves represent components $t_1$ and $t_2$ of the GMM approximation of the activation histogram. The shaded area denotes range of $\mu_{t} \pm \sigma^2_t$ and colored dots demonstrate possible encodings for exemplary values of $b_v$. Legend associates dot color with the value of the encoding bits under corresponding GMM given by $[t_1, t_2]$.}
    \label{fig:encoding}
\end{figure}

Final binary perceptual code $\mathbf{e}_\mathbf{I}$ of image $\mathbf{I}$ is given by concatenation of encoded value representations $\mathbf{e}_v$ of all relevant activation values $v$. The illustration of the affiliation computation is depicted in Figure \ref{fig:encoding}. 

\subsection{Atlas creation and prediction basis retrieval} \label{sec:atlasCreation}
Encoding entire training dataset (and possibly other datasets) results in encoded atlas representation $\mathbf{P}^{(c)} \in \mathbb{N}^{K \times N}$ for class $c$ where $K$ is the number of encoded images of the given class and $N = |\mathbf{e}_\mathbf{I}| = MT$ denotes the length of encoded vector $\mathbf{e}_\mathbf{I}$ of each image.

Comparison of two encoded image representations $\mathbf{e}_{I_1}$ and $\mathbf{e}_{I_2}$ is done via calculation of their Hamming distance. Given that $\mathbf{e}_{\mathbf{I}_1}$ and $\mathbf{e}_{\mathbf{I}_2}$ are binary vectors, this is easily implemented as element-wise (or even bit-wise if correctly compressed) XOR operation. Hamming distance can be weighted using vector $\mathbf{w}^{(c)} \in \mathbb{R}^N$ having $i$-th element defined by:

\begin{equation}\label{eq:encodingWeight}
w^{(c)}_{i} = \sum_{j=0}^{K} P^{(c)}_{ji}
\end{equation}

%The final prediction basis consists of $n$ atlas images with the smallest distances between their and the unknown test image encoded representations.

The final prediction basis is given by $k$ nearest neighbors from the respective atlas, as measured by the Hamming distance of their corresponding binary neural activation codes.

%\subsection{Running Example}

%To illustrate this step, consider the first 3D convolutional layer of a typical 3D CNN used for volumetric image processing, followed by a 3D max-pooling layer. The convolutional layer consists of 8 filters of size $3 \times 3 \times 3$ and the max-pooling layer with window size $2 \times 2 \times 2$ and stride $2$. Input data of the network are $90 \times 90 \times 90$ volumetric images (e.g. MRI scans). Considering $1000$ input images of a single class, the output feature maps of the convolutional layer have a size of $[1000,5832000‬]$ and down-sampled feature maps after max-pooling layer have a size of $[1000,729000]$. Deciding to generate histograms with $60$ bins from the max-pooling layer results in $729000$ histograms, each binning the $1000$ activation values from the input images into $60$ bins.

%The approximation of all $729000$ histograms results in array of size $[729000,8]$.

%Each of the $729000$ histograms is encoded into 2 bits resulting in $1458000$ bits of encoding information from the entire max-pooling layer.

\section{Evaluation}
A proof-of-concept evaluation on a small volumetric image dataset 3D MNIST \footnote{https://www.kaggle.com/daavoo/3d-mnist} has been reported in a prior work on this method~\cite{stano2020explainable}. There, our primary goal was to visually evaluate the retrieved prediction basis for both datasets and compare both ends of the prediction basis, i.e. the most activation-wise similar and dissimilar images, to the unknown test image. The secondary goal was to evaluate the memory efficiency of the encoded atlas representation. 

Here, we evaluate our method on a custom neuroimaging dataset that was systematically constructed by selection, labelling, and augmentation by distortion of samples from an existing neuroimaging dataset, with a particular objective in mind, as described below in Section \ref{sub::dataset}.

\subsection{Performance Quantifier}

To enable quantitative evaluation of our explainability method without a need for expensive user studies on practicing medical experts, we devised an evaluation pipeline based on an assumption, that given an unknown test image $\mathbf{I}$ of class $c'$ and prediction basis $\mathbb{B}$, the accuracy of our method is proportional to
\begin{enumerate}
    \item the fraction of prediction basis images $\mathbf{B}_{i}$ from class $c'$ of full prediction basis, and
    \item the position $i$ in the prediction basis, meaning that $\mathbf{B}_{i}$ has a higher impact on final accuracy than $\mathbf{B}_{i+1}$,
\end{enumerate}
as $\mathbf{B}_1$ is the most activation-wise similar atlas image to the unknown image $\mathbf{I}$. In this context, $c'$ is an intra-class of class $c$, e.g. Alzheimer disease positive (class) with frontal ventricle atrophy (intra-class). We define our "prediction basis accuracy" metric $P_{acc}$ by:

\begin{equation}\label{eq:basisMetric}
P_{acc}(\mathbf{I},\mathbb{B}) = \sum_{i=1}^{k} {\frac{sim(\mathbf{I}, \mathbf{B}_i)}{2^i}}
\end{equation}

where $sim(\mathbf{I}, \mathbf{B}_i)$ results in $1$ if both $\textbf{I}$ and $\mathbf{B}_i$ are in the same intra-class $c'$ and $0$ in all  other cases.

\subsection{Dataset Construction}
\label{sub::dataset}
Finding or creating a neuroimaging dataset with labeled natural intra-classes (like the one in the example above) is a difficult task requiring the aid of a domain expert. 

For this reason, we create multiple synthesized datasets with artificial intra-classes. We picked the TADPOLE grand-challenge dataset \footnote{https://tadpole.grand-challenge.org} for Alzheimer's disease classification as the basis for these artificially synthesized datasets.

\begin{figure}[h]
    \centering
    \includegraphics[width=\columnwidth]{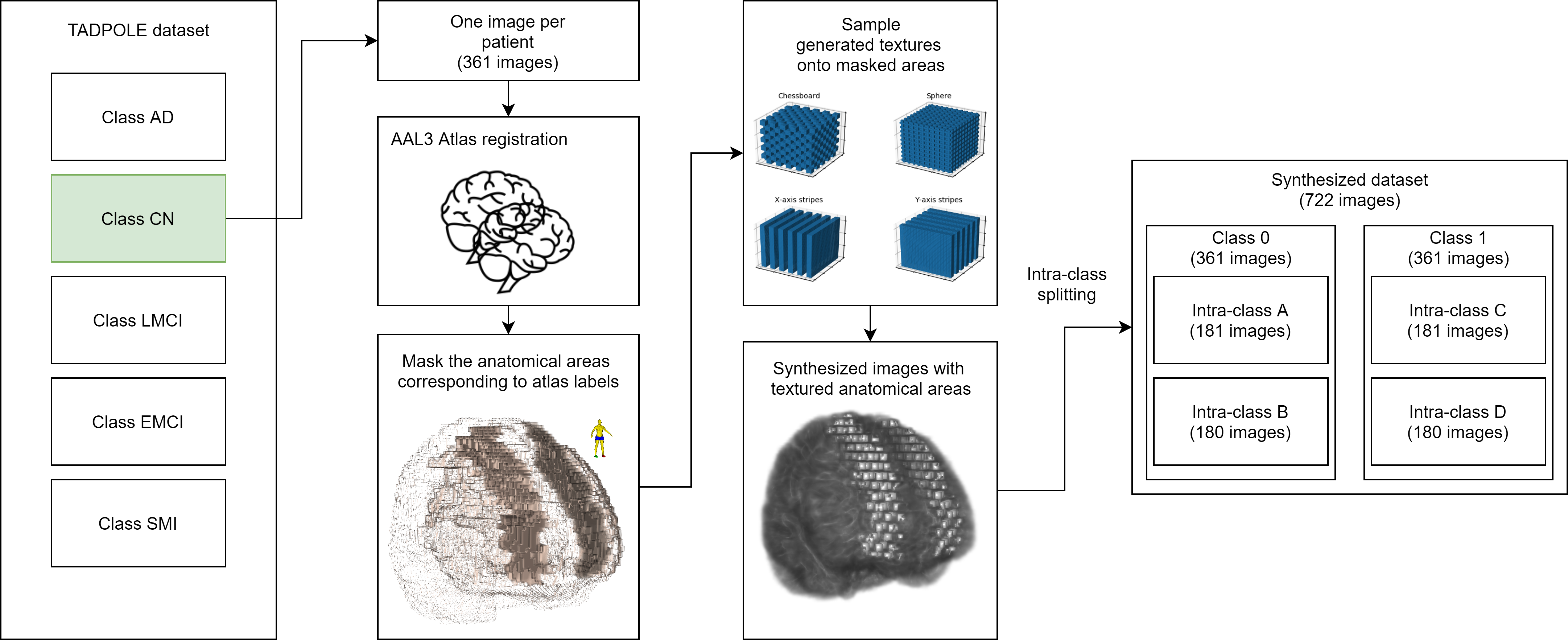}
    \caption{Creating synthesized dataset - an example for two intra-classes per one class.}
    \label{fig:syntesizedDatasetCreation}
\end{figure}

The intra-classes $C'$ were then introduced as disjoint sets of pattern-textured anatomical brain areas with each set representing a single intra-class. Each image from the base TADPOLE dataset was textured with a unique synthetic volumetric texture pattern (see Fig. \ref{fig:syntesizedDatasetCreation}) according to its synthetic intra-class label, to produce the resulting synthesized dataset. A model was later trained to classify two classes, each of which contained the same number of distinct intra-classes.

We selected one image per patient from the CN (cognitively normal) class of the base dataset for a total of $361$ images per class of the resulting synthesized dataset. Multiple images of the same patient in one class of the resulting dataset could be deemed as a natural intra-class and thus spoil the isolated nature of our experiments.

To make the texturing of the anatomical areas possible, each chosen image registered with the AAL3 atlas \cite{rolls2020automated}, using the ANTs toolkit \cite{klein2009elastix}. The process of creating a synthesized textured dataset is depicted in Fig. \ref{fig:syntesizedDatasetCreation}.

\subsection{Experimental Configuration}

The topology (architecture) of the analyzed 3D CNN model is depicted in detail in Table \ref{tab:architecture}. The model was trained to minimize categorical cross-entropy loss function using stochastic gradient descent.

\begin{table}[h]
\centering
\resizebox{\columnwidth}{!}{%
\begin{tabular}{|l|l|l|l|l|l|l|l|l|}
\hline
Layer    & \begin{tabular}[c]{@{}l@{}}Filters/\\ Units\end{tabular} & \multicolumn{1}{c|}{\begin{tabular}[c]{@{}c@{}}Kernel \\ Size\end{tabular}} & \multicolumn{1}{c|}{Stride} & Activation & \begin{tabular}[c]{@{}l@{}}Batch \\ Normalization\end{tabular} & \begin{tabular}[c]{@{}l@{}}Weight \\ regularization\end{tabular} & \begin{tabular}[c]{@{}l@{}}Bias \\ regularization\end{tabular} & Dropout \\ \hline
Conv3D 1 & 8                                                        & 5                                                                           & 1                           & LeakyRelu  & Yes                                                            & 0.001                                                            & 0.001                                                          & No      \\ \hline
Conv3D 2 & 8                                                        & 5                                                                           & 2                           & LeakyRelu  & Yes                                                            & 0.001                                                            & 0.001                                                          & No      \\ \hline
Conv3D 3 & 8                                                        & 5                                                                           & 1                           & LeakyRelu  & Yes                                                            & 0.001                                                            & 0.001                                                          & No      \\ \hline
Conv3D 4 & 8                                                        & 5                                                                           & 2                           & LeakyRelu  & Yes                                                            & 0.001                                                            & 0.001                                                          & No      \\ \hline
Conv3D 5 & 8                                                        & 5                                                                           & 1                           & LeakyRelu  & Yes                                                            & 0.001                                                            & 0.001                                                          & No      \\ \hline
Conv3D 6 & 8                                                        & 5                                                                           & 2                           & LeakyRelu  & Yes                                                            & 0.001                                                            & 0.001                                                          & No      \\ \hline
Dense 1  & 128                                                      & N/A                                                                         & N/A                         & LeakyRelu  & No                                                             & No                                                               & No                                                             & 0.3     \\ \hline
Dense\_2 & 2                                                        & N/A                                                                         & N/A                         & Softmax    & No                                                             & No                                                               & No                                                             & No      \\ \hline
\end{tabular}}
\caption{Topology of the deep 3D convolutional neural network used for validation of our explainability method.}
\label{tab:architecture}
\end{table}

To provide further insight into performance of our proposed method, we visualize the encoded atlas representations for a given class, by applying PCA for dimensionality reduction to the encoded atlas representation $\mathbf{P}^{(c)}$ (as defined in Section \ref{sec:atlasCreation}), resulting in transformed representation $\mathbf{P}_{r}^{(c)} \in \mathbb{N}^{k \times 2}$.

\begin{table}[h]
\subfloat[Scenario 1]{
\resizebox{\columnwidth}{!}{%
\begin{tabular}{|l|l|l|l|l|}
\hline
\multirow{2}{*}{Class} & \multirow{2}{*}{Intra-class} & \multicolumn{2}{l|}{Anatomical area}         & \multirow{2}{*}{\begin{tabular}[c]{@{}l@{}}Texturing \\ pattern\end{tabular}} \\ \cline{3-4}
                       &                              & Names                          & AAL3 Labels &                                                                               \\ \hline
\multirow{2}{*}{0}     & A                            & FrontalSup2L, FrontalSup2R     & 3, 4        & Chessboard                                                                    \\ \cline{2-5} 
                       & B                            & TempSupL, TempSupR             & 85, 86      & Sphere                                                                        \\ \hline
\multirow{2}{*}{1}     & C                            & SuppMotorAreaL, SuppMotorAreaR & 15, 16      & Sphere                                                                        \\ \cline{2-5} 
                       & D                            & PrecuneusL,  PrecuneusR        & 71, 72      & Chessboard                                                                    \\ \hline
\end{tabular}}}

\vfill

\subfloat[Scenario 2]{
\resizebox{\columnwidth}{!}{%
\begin{tabular}{|l|l|l|l|l|}
\hline
\multirow{2}{*}{Class} & \multirow{2}{*}{Intra-class} & \multicolumn{2}{l|}{Anatomical area}         & \multirow{2}{*}{\begin{tabular}[c]{@{}l@{}}Texturing \\ pattern\end{tabular}} \\ \cline{3-4}
                       &                              & Names                          & AAL3 Labels &                                                                               \\ \hline
\multirow{4}{*}{0}     & A                            & FrontalSup2L, FrontalSup2R     & 3, 4        & Chessboard                                                                    \\ \cline{2-5} 
                       & B                            & TempSupL, TempSupR             & 85, 86      & Sphere                                                                        \\ \cline{2-5} 
                       & C                            & InsulaL, InsulaR               & 33, 34      & X-axis stripes                                                                \\ \cline{2-5} 
                       & D                            & HippocampusL, HippocampusR     & 41, 42      & Y-axis stripes                                                                \\ \hline
\multirow{4}{*}{1}     & E                            & SuppMotorAreaL, SuppMotorAreaR & 89, 90      & Chessboard                                                                    \\ \cline{2-5} 
                       & F                            & PrecuneusL,  PrecuneusR        & 59, 60      & Sphere                                                                        \\ \cline{2-5} 
                       & G                            & Cerebellum45R, Cerebellum6L    & 102, 103    & X-axis stripes                                                                \\ \cline{2-5} 
                       & H                            & OccipitalInfL, OccipitalInfR   & 57, 58      & Y-axis stripes                                                                \\ \hline
\end{tabular}}}
\caption{Details of the artificial intra-classes.}
\label{tab:intraClasses}
\end{table}

We designed two evaluation scenarios and executed them on two different artificially synthesized datasets. The details of the intra-class composition of these two datasets are depicted in Table \ref{tab:intraClasses}.

\subsection{Results}

The average prediction basis accuracy (Eq. \ref{eq:basisMetric}) achieved by our explainability method in the experiments w.r.t. the convolutional layer used for creation of neuron activity histograms (\ref{sub::histo}) is shown in Table \ref{tab:resultAcc}.

\begin{table}[h]
\centering
\resizebox{\columnwidth}{!}{%
\begin{tabular}{|l|l|l|l|l|l|l|}
\hline
\multirow{2}{*}{Scenario} & \multirow{2}{*}{\begin{tabular}[c]{@{}l@{}}Model test \\ accuracy\end{tabular}} & \multicolumn{2}{l|}{\begin{tabular}[c]{@{}l@{}}Number of \\ test images\end{tabular}} & \multirow{2}{*}{\begin{tabular}[c]{@{}l@{}}Network layer\\ analysed\end{tabular}} & \multicolumn{2}{l|}{\begin{tabular}[c]{@{}l@{}}Average prediction basis\\ accuracy score\end{tabular}} \\ \cline{3-4} \cline{6-7} 
                          &                                                                                 & Class 0                                   & Class 1                                   &                                                                                   & Class 0                                            & Class 1                                           \\ \hline
\multirow{3}{*}{1}        & \multirow{3}{*}{0.9931}                                                         & \multirow{3}{*}{73}                       & \multirow{3}{*}{72}                       & Conv3D 2                                                                          & 0.9423 $\pm$ 0.310                                 & 0.8865 $\pm$ 0.218                                \\ \cline{5-7} 
                          &                                                                                 &                                           &                                           & Conv3D 4                                                                          & 0.9151 $\pm$ 0.305                                 & 0.8815 $\pm$ 0.172                                \\ \cline{5-7} 
                          &                                                                                 &                                           &                                           & Conv3D 6                                                                          & 0.9358 $\pm$ 0.308                                 & 0.8720 $\pm$ 0.203                                \\ \hline
\multirow{3}{*}{2}        & \multirow{3}{*}{0.9931}                                                         & \multirow{3}{*}{73}                       & \multirow{3}{*}{72}                       & Conv3D 2                                                                          & 0.4816 $\pm$ 0.159                                 & 0.4154 $\pm$ 0.293                                \\ \cline{5-7} 
                          &                                                                                 &                                           &                                           & Conv3D 4                                                                          & 0.6485 $\pm$ 0.182                                 & 0.5597 $\pm$ 0.293                                \\ \cline{5-7} 
                          &                                                                                 &                                           &                                           & Conv3D 6                                                                          & 0.7149 $\pm$ 0.147                                 & 0.6937 $\pm$ 0.256                                \\ \hline
\end{tabular}}
\caption{Results at different layers and scenarios.}
\label{tab:resultAcc}
\end{table}

\begin{figure}[hbt!]
    \centering
    \subfloat[Prediction basis for test image from class 0]{\includegraphics[width=\columnwidth]{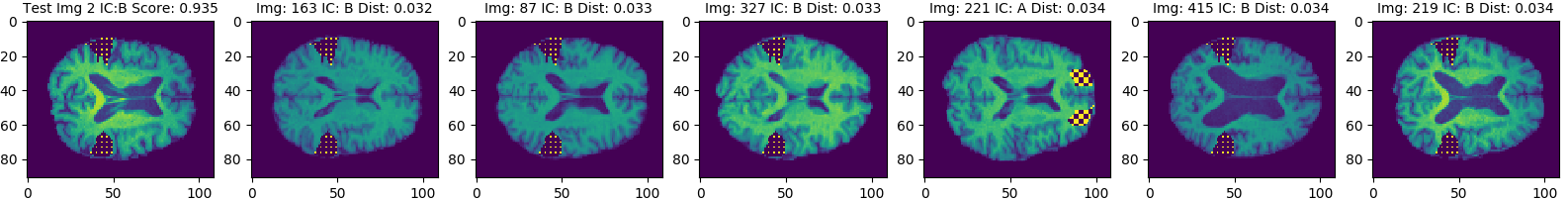}}
    \vfill
    \subfloat[Prediction basis for test image from class 1]{\includegraphics[width=\columnwidth]{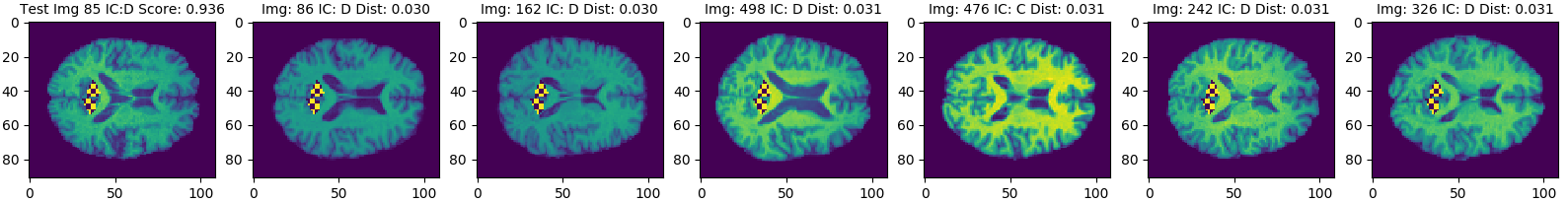}}
    \vfill
%    \subfloat[Encoded atlas for class 0]{\label{subfig:atlas0Scenario1}\includesvg[height=130pt]{newResults/scenario_1_layer_3_class_0_cluster}}
     \subfloat[Encoded atlas for class 0]{\label{subfig:atlas0Scenario1}\includegraphics[height=130pt]{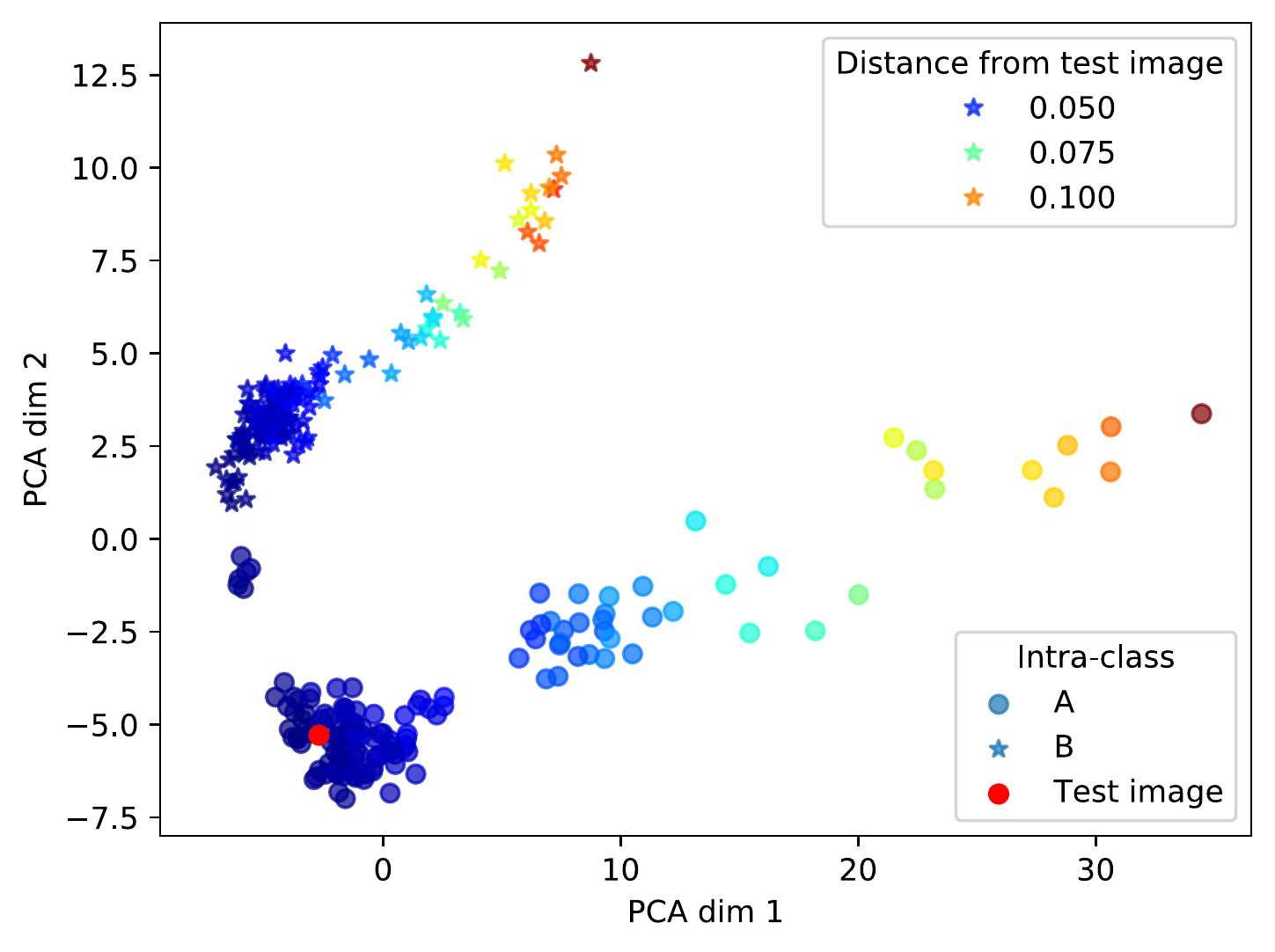}}
    \hfill
%    \subfloat[Encoded atlas for class 1]{\label{subfig:atlas1Scenario1}\includesvg[height=130pt]{newResults/scenario_1_layer_3_class_1_cluster}}
   \subfloat[Encoded atlas for class 1]{\label{subfig:atlas1Scenario1}\includegraphics[height=130pt]{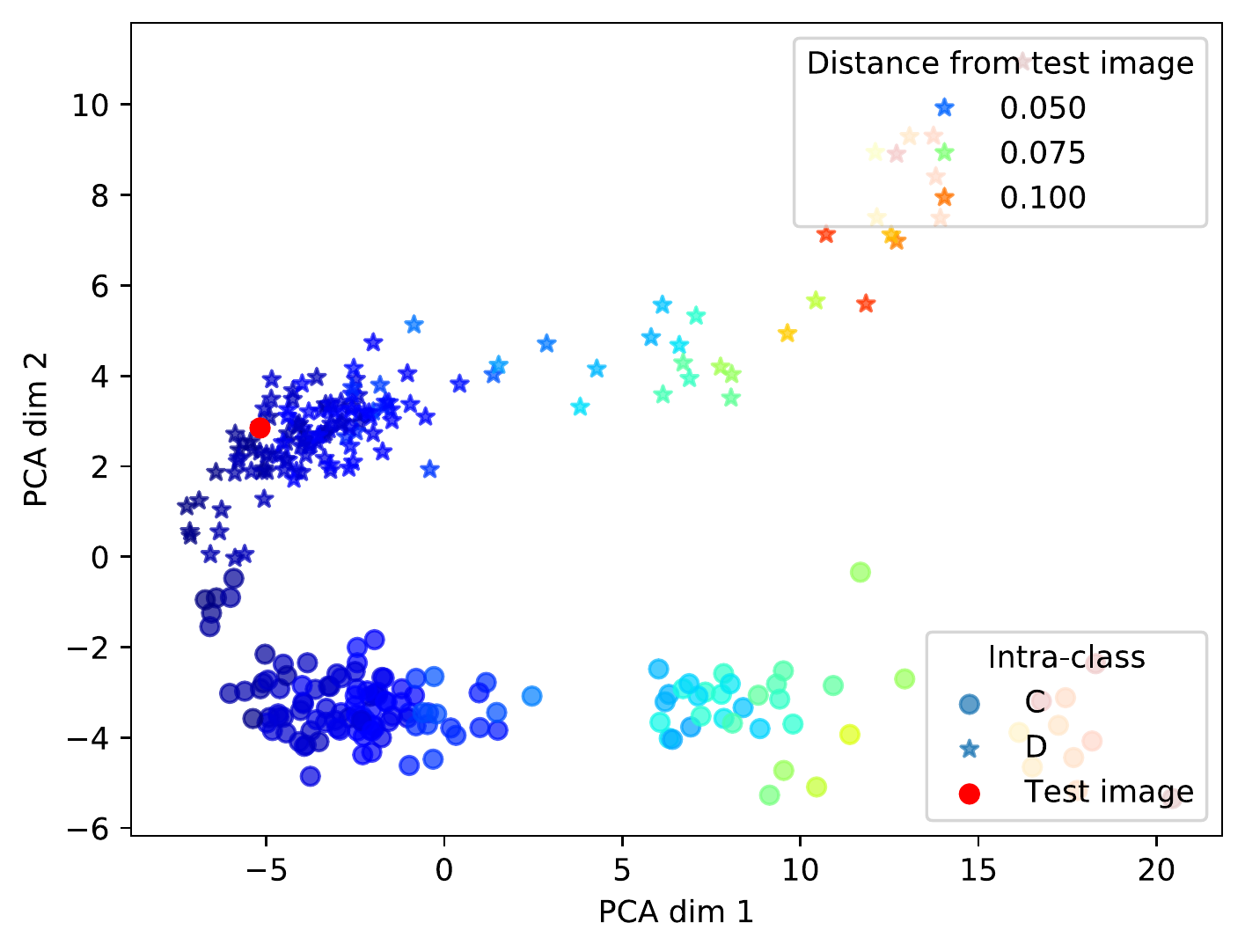}}
    \caption{Prediction basis and transformed encoded atlas representations for scenario 1.}
    \label{fig:resultScenario1}
\end{figure}

\begin{figure}[hbt!]
    \centering
    \subfloat[Prediction basis for test image from class 0]{\includegraphics[width=\columnwidth]{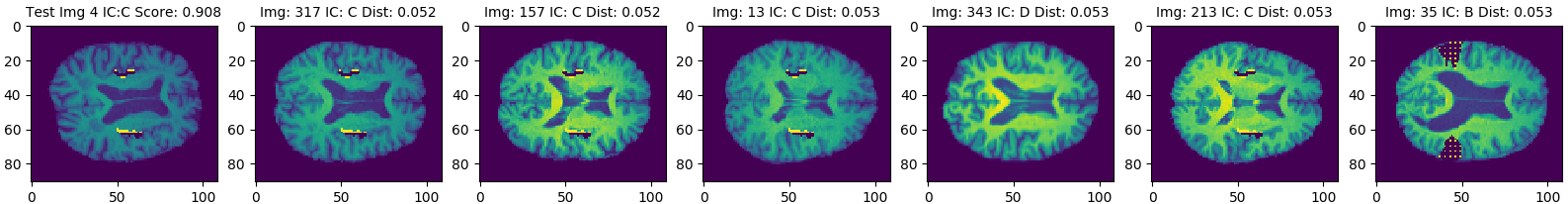}}
    \vfill
    \subfloat[Prediction basis for test image from class 1]{\includegraphics[width=\columnwidth]{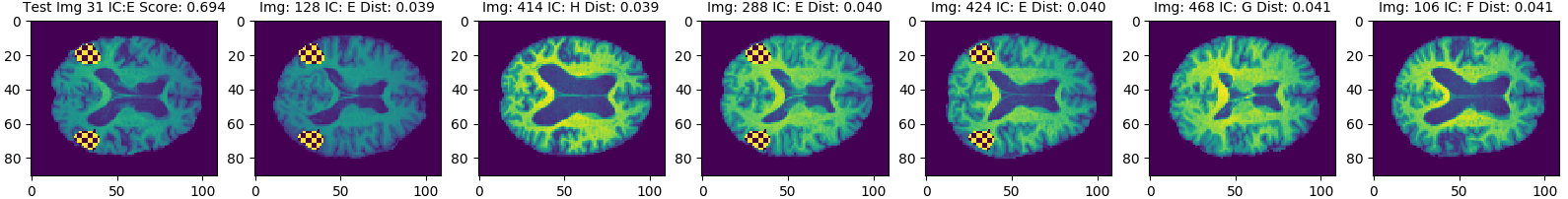}}
    \vfill
%    \subfloat[Encoded atlas for class 0]{\label{subfig:atlas0Scenario2}\includesvg[height=127pt]{newResults/scenario_2_layer_3_class_0_cluster}}
  \subfloat[Encoded atlas for class 0]{\label{subfig:atlas0Scenario2}\includegraphics[height=127pt]{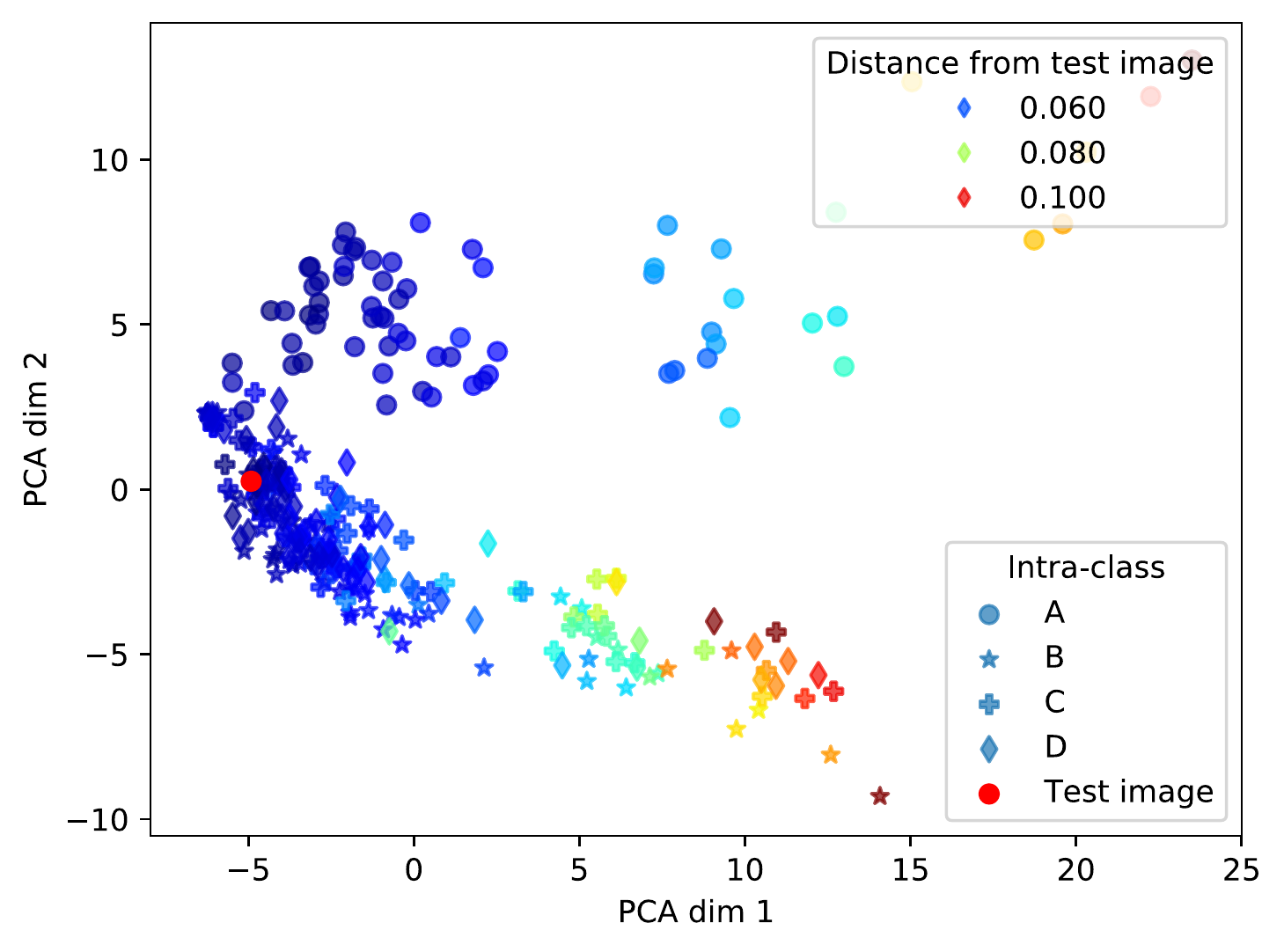}}
    \hfill 
%    \subfloat[Encoded atlas for class 1]{\includesvg[height=127pt]{newResults/scenario_2_layer_3_class_1_cluster}}
    \subfloat[Encoded atlas for class 1]{\includegraphics[height=127pt]{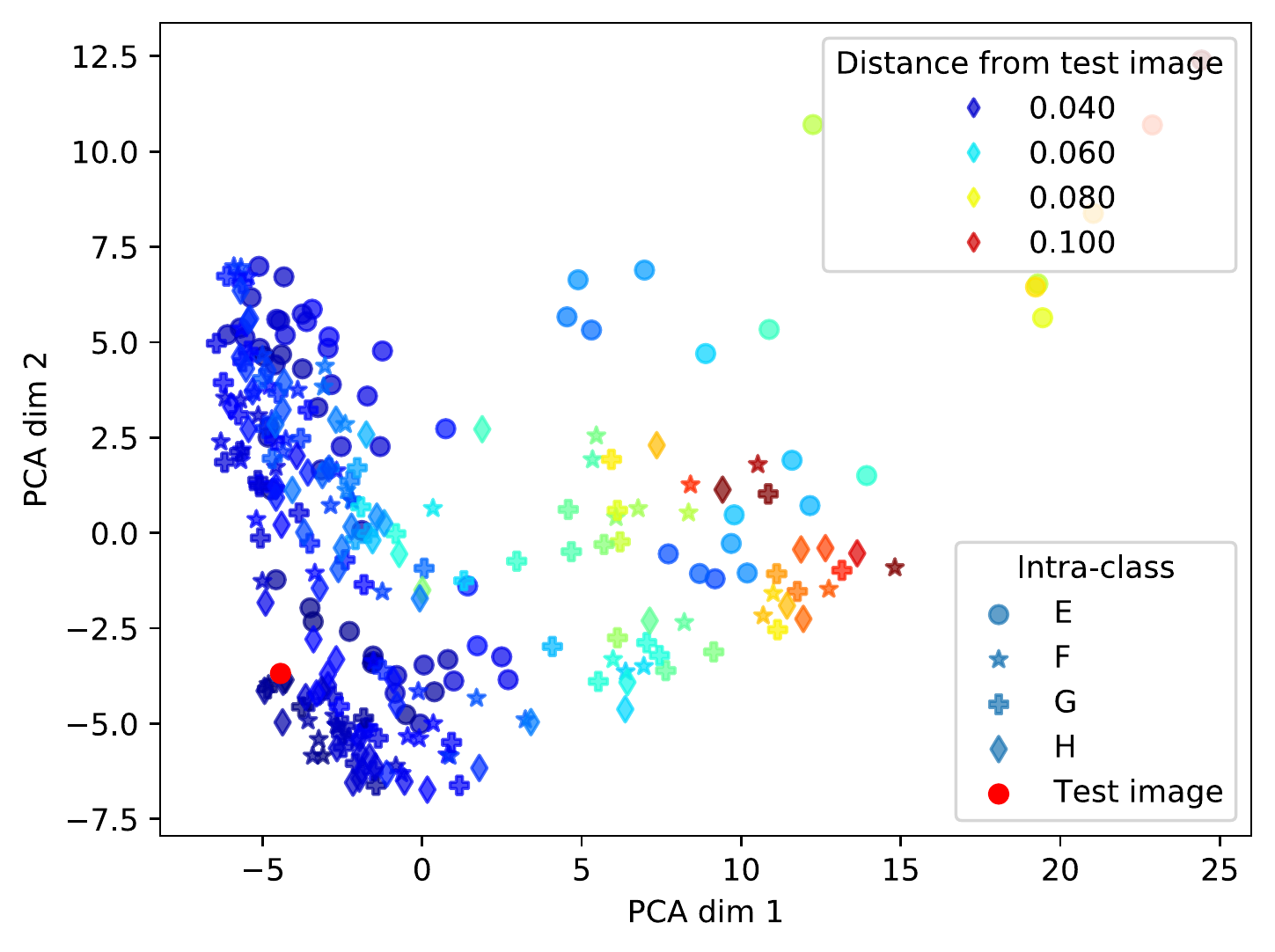}}
    \caption{Prediction basis and transformed encoded atlas representations for scenario 2.}
    \label{fig:resultScenario2}
\end{figure}

Results suggest, that the activity analysis of deeper network layers leads to higher average prediction basis accuracy. Visual examples of the retrieved prediction basis (in a form of a central axial slice of the full volume) and visualizations of the PCA transformed encoded atlas representations for the deepest analyzed convolutional layer (\textit{Conv3D 6}) and both scenarios are shown in Fig. \ref{fig:resultScenario1} and Fig. \ref{fig:resultScenario2}.

In the case of scenario 1, an intra-class-wise clustering of the PCA transformed encoded atlas entries is visible in Fig. \ref{subfig:atlas0Scenario1} and Fig. \ref{subfig:atlas1Scenario1} which corresponds to the high accuracy of our prediction basis retrieval in this scenario.

On the other hand, in scenario 2 where the average achieved prediction basis accuracy was lower, the clustering of PCA transformed encoded atlas is also much less clear with mostly overlapping clusters except for intra-class \textit{A} in Fig. \ref{subfig:atlas0Scenario2}.

\subsection{Verifying Detected Prediction Strategies}
% Layer-wise Relevance Propagation
% Additional Verification
% External Confirmation / Validation
% Prediction Strategy Verification

\begin{figure}[h]
    \centering
    \subfloat[Intra-class A]{\label{subfig:lrp1a}\includegraphics[width=0.25\columnwidth]{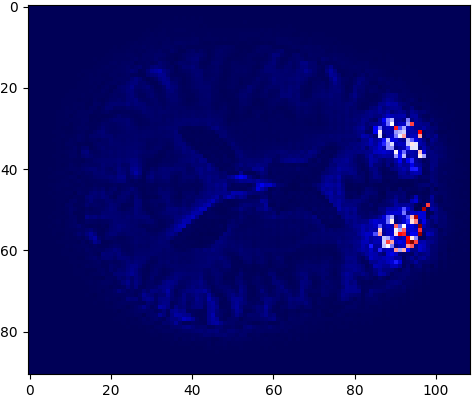}}
    \hfill
    \subfloat[Intra-class B]{\label{subfig:lrp1b}\includegraphics[width=0.25\columnwidth]{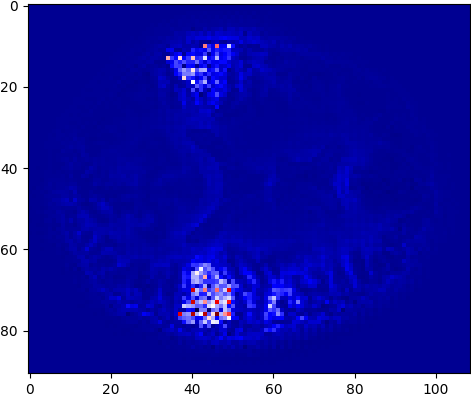}}
    \hfill
    \subfloat[Intra-class C]{\label{subfig:lrp1c}\includegraphics[width=0.25\columnwidth]{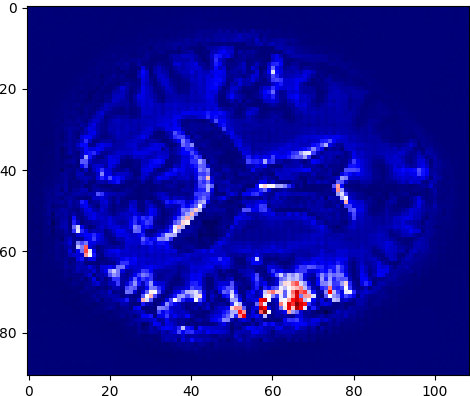}}
    \hfill
    \subfloat[Intra-class D]{\label{subfig:lrp1d}\includegraphics[width=0.25\columnwidth]{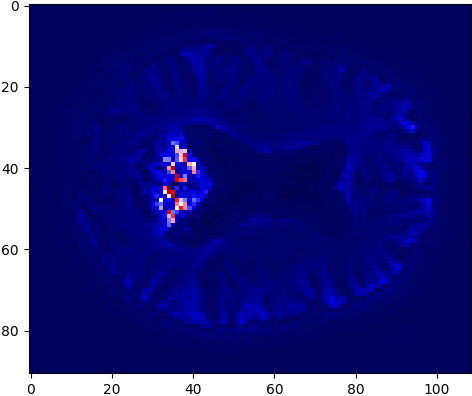}}
    \caption{LRP analysis of intra-classes for scenario 1.}
    \label{fig:lrp1}
\end{figure}

To further interpret and complement these results, we incorporate the LRP (layer-wise relevance propagation) method \cite{bach2015pixel} into the classification process of our neural network model. We mainly focused on the association between the relevancy of voxels in the textured areas of the artificial intra-classes and the clustering properties of the corresponding intra-classes in the PCA transformed encoded atlases. The results of the LRP analysis of one sample training image per each artificial intra-class used in the experimental scenario 1 are depicted in Fig. \ref{fig:lrp1}

\begin{figure}[]
    \centering
    \subfloat[Intra-class A]{\label{subfig:lrp2a}\includegraphics[width=0.25\columnwidth]{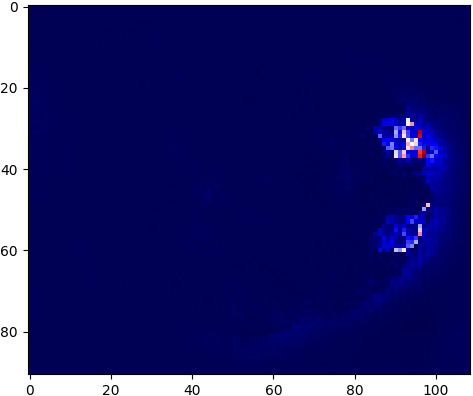}}
    \hfill
    \subfloat[Intra-class B]{\label{subfig:lrp2b}\includegraphics[width=0.25\columnwidth]{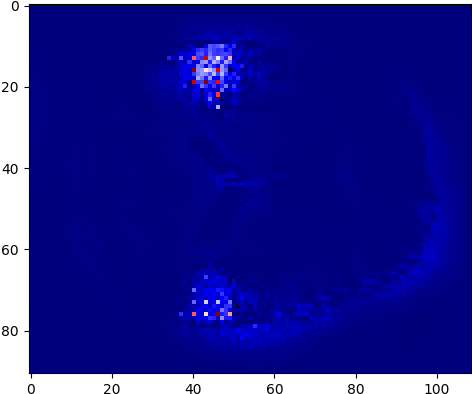}}
    \hfill
    \subfloat[Intra-class C]{\label{subfig:lrp2c}\includegraphics[width=0.25\columnwidth]{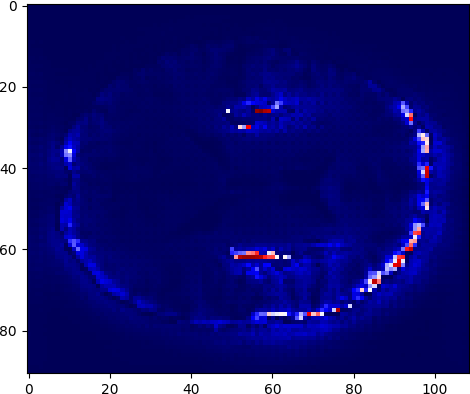}}
    \hfill
    \subfloat[Intra-class D]{\label{subfig:lrp2d}\includegraphics[width=0.25\columnwidth]{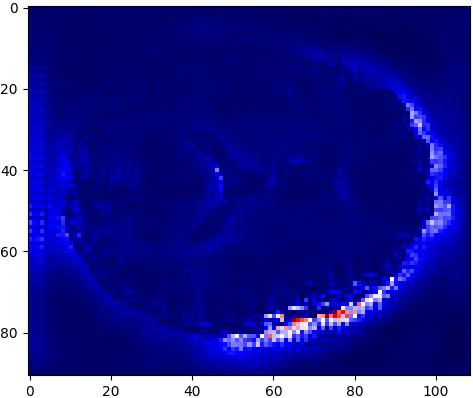}}
    \vfill
    \subfloat[Intra-class E]{\label{subfig:lrp2e}\includegraphics[width=0.25\columnwidth]{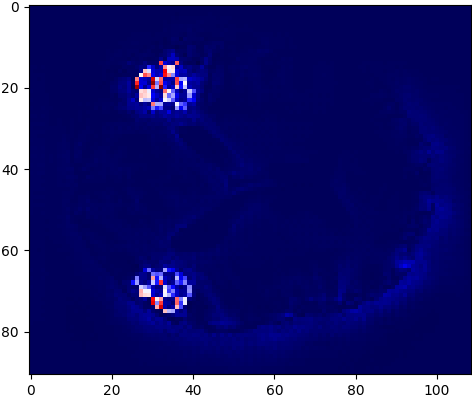}}
    \hfill
    \subfloat[Intra-class F]{\label{subfig:lrp2f}\includegraphics[width=0.25\columnwidth]{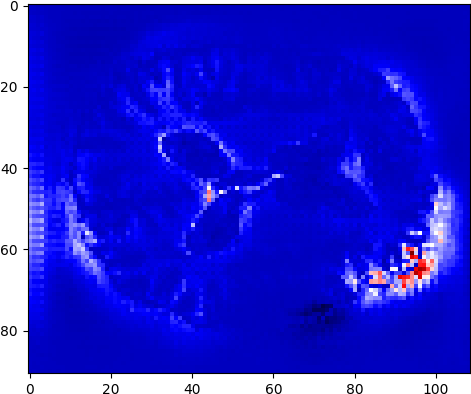}}
    \hfill
    \subfloat[Intra-class G]{\label{subfig:lrp2g}\includegraphics[width=0.25\columnwidth]{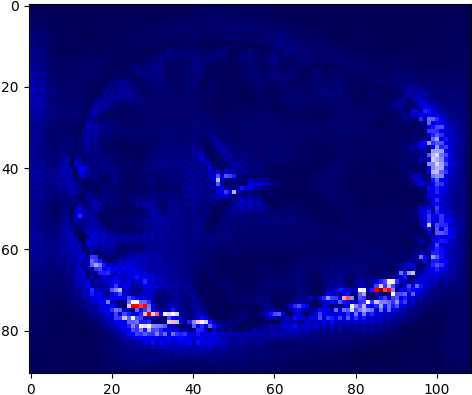}}
    \hfill
    \subfloat[Intra-class H]{\label{subfig:lrp2h}\includegraphics[width=0.25\columnwidth]{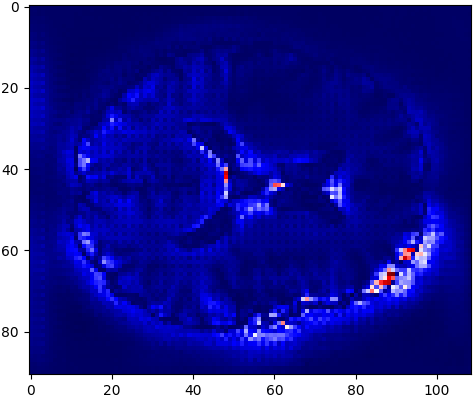}}
    \caption{LRP analysis of intra-classes for scenario 2.}
    \label{fig:lrp2}
\end{figure}

It is possible to observe a strong relevancy of the textured areas for both sample intra-class images of class $0$ in Fig. \ref{subfig:lrp1a} and \ref{subfig:lrp1b}. In the case of class $1$, the relevancy only matches textured areas of intra-class \textit{D} image (Fig. \ref{subfig:lrp1d}) and is more focused on the natural edges in the intra-class \textit{C} image (Fig. \ref{subfig:lrp1c}). We also similarly analyzed the intra-classes of the dataset used for the experimental scenario 2 with results depicted in Fig. \ref{fig:lrp2}.

A similar observation as in scenario 1 can be made here. Regardless of class, certain intra-class texturing areas (Fig. \ref{subfig:lrp2a}, \ref{subfig:lrp2b} and \ref{subfig:lrp2e}) matches the highly relevant voxels while others do not. An association valid for both scenarios is that the intra-classes which have their textured areas matched with high relevancy are more distinguishably clustered in the visualizations of their corresponding PCA reduced encoded atlases. 

Based on the LRP results and given that clustering of a single prediction explainability heatmaps (such as the ones produced by LRP) was already used for a successful detection of model's distinct prediction strategies \cite{lapuschkin2019unmasking}, we assume that clustering the LRP heatmaps in our experiments would produce very similar results as visualizations of our PCA transformed encoded atlases. Under this assumption we deem our method equally able to detect the model's distinct prediction strategies.

\section{Conclusion}

We devised a novel explainability method for deep convolutional neural networks based on a novel binary representation of neural perceptual code. The main goal of the devised method was to produce a prediction basis to support the explanation of particular classification decision to the operating personnel. This is achieved by retrieving $k$ most activation-wise similar annotated (training) images to the test image.

We deemed our method to be most applicable in the domain of medical imaging and specifically neuroimaging. Therefore we test it on two artificially synthesized datasets based on the ADNI TADPOLE neuroimaging dataset. The usage of synthesized datasets with artificially textured anatomical areas allowed us to quantitatively evaluate the performance of the proposed method and also made the visual examples of the prediction basis readable by non-domain experts.

Besides providing prediction basis for single predictions, we also attempted to visualize our custom created atlas composed of all binary encoded training images of a given class, i.e. the whole pool from which each prediction basis is drawn.

It is visible from the provided prediction basis examples that the majority of the prediction basis images contains the same texturing pattern as the unknown test image. This observation is complemented by the encoded atlas visualizations in which natural clusters had been formed corresponding to the differently textured intra-classes of the synthesized dataset.

In the case of the second synthesized dataset where the clear natural clustering of the encoded atlas is less prevalent we employed the layer-wise relevance propagation (LRP) technique to explain this behavior. The LRP assigned relevancy to individual voxels of the input image based on the hidden neural activations. It is observable that clusters in the encoded atlas visualization were well-defined if the high relevancy matched the artificially textured areas and were overlapping. Conversely, relevancy for intra-classes with overlapping and malformed clusters was scattered over other brain regions. 

Therefore less accurate prediction basis in some cases is not caused by shortcomings of our explainability method but by similar neural activation patterns for different intra-classes learned during the model's training process. In fact, analyzing the encoded atlas visualizations might be useful for the detection of an undesired over-generalization of the trained model.

\bibliography{mybibfile}

\end{document}